%% file: feature-paper.tex
\documentclass[11pt,draftcls,onecolumn,journal]{IEEEtran}

\usepackage{amsmath,amsfonts, amsthm}
\usepackage{mathtools}
\usepackage{bbm}
\usepackage{algorithmic}
\usepackage{algorithm}
\usepackage{array}
\usepackage[caption=false,font=normalsize,labelfont=sf,textfont=sf]{subfig}
\usepackage{textcomp}
\usepackage{stfloats}
\usepackage{url}
\usepackage{graphicx}
\usepackage{cite}
\usepackage{xcolor}
\usepackage{hyperref}
\hypersetup{colorlinks=true, linkcolor=blue, citecolor=blue, urlcolor=blue}
\usepackage{booktabs}
\usepackage{pifont}
\usepackage{cleveref}
\usepackage{tikz}
\usetikzlibrary{arrows.meta,calc,positioning}

\theoremstyle{definition}
\newtheorem*{defn}{Definition}
\theoremstyle{theorem}

\newtheorem{thm}{Theorem}
\tikzset{
  manifold/.style={fill=blue!7, draw=blue!55, line width=0.9pt},
  rim/.style={draw=blue!55, line width=0.9pt},
  tangent/.style={fill=orange!12, draw=orange!75, line width=0.7pt},
  wpt/.style={circle, fill=black, inner sep=1.35pt},
  gstep/.style={-{Stealth[length=2.4mm,width=1.8mm]}, line width=0.95pt},
  pull/.style={-{Stealth[length=2.4mm,width=1.8mm]}, line width=0.95pt, red!65!black},
  geo/.style={-{Stealth[length=2.4mm,width=1.8mm]}, line width=1.1pt, teal!55!black},
  aux/.style={dashed, gray!75, line width=0.6pt},
  lbl/.style={font=\small},
  slbl/.style={font=\footnotesize},
  rimhidden/.style      = {draw=blue!55, line width=0.9pt, densely dotted},
  geosolid/.style       = {line width=1.3pt, teal!55!black},
  geohidden/.style      = {line width=1.3pt, teal!55!black, densely dotted},
  geohiddenarrow/.style = {line width=1.3pt, teal!55!black, densely dotted,                         -{Stealth[length=2.4mm,width=1.8mm]}},
}
\newcommand{\blobpath}{plot[smooth cycle, tension=0.8] coordinates
  {(-3.0,0.15) (-1.4,-0.95) (0.9,-1.15) (2.8,-0.45) (3.35,0.85)
   (1.7,1.75) (-0.4,1.30) (-2.3,1.65)}}

\newcommand{\drawblob}{%
  \draw[manifold, even odd rule]
    \blobpath
    (1.6,0.5) arc(120:60:1 and 1.25) arc(-60:-120:1 and 1.25);
  \draw[rim] (1.6,0.5) arc(-120:-130:1 and 1.25)
             ( 2.6,0.5) arc( -60: -50:1 and 1.25);
}

\DeclareMathOperator*{\argmax}{arg\,max}

\input{commons-utils}
\begin{document}
\title{\thetitle}

\author{
    \IEEEauthorblockN{Fabio Brau, Giorgio Piras, Maura Pintor,~\IEEEmembership{Member,~IEEE} and Battista Biggio,~\IEEEmembership{Fellow,~IEEE}}\\
    \IEEEauthorblockA{
        Department of Electrical and Electronic Engineering,\\
        University of Cagliari, Cagliari, Italy\\
    }
    \thanks{Emails: \{fabio.brau; giorgio.piras; maura.pintor; battista.biggio\}@unica.it}
}
\markboth{Submitted to IEEE Signal Processing Magazine}%
{Brau \MakeLowercase{\textit{et al.}}: Lipschitz Thinking}

\maketitle

The Lipschitz property of a deep neural network provides a direct measure of its sensitivity to input perturbations and, when explicitly controlled, offers a principled way to limit the propagation of errors and improve robustness. Over the past decade, Lipschitz-bounded layers have been incorporated into increasingly expressive and high-performing deep models, narrowing the gap between empirical robustness and formal, by-design guarantees of stability. This article introduces the fundamental concepts underlying Lipschitz-bounded neural networks, explaining the principles behind Lipschitz-constrained layers, the mechanisms used to enforce their bounds, and how they yield robustness certificates at the cost of a single forward pass. The tutorial concludes by discussing emerging and open directions, highlighting Lipschitz control as a general framework for offering guaranteed, by-design stability.


\section*{Motivation}
\label{sec:intro}


\IEEEPARstart{L}{earned} models have become ordinary building blocks of the signal processing chains. Deep networks operate at the perception stage of several systems, e.g., autonomous vehicles~\cite{kendall2019learning} and radiologic assistants for reading medical images~\cite{liang2024medical}.
The uncomfortable discovery of the past decade, \textit{adversarial examples}~\cite{biggio2013evasion, szegedy2014intriguing}, reveals how brittle Deep Neural Network (DNN)-based pipelines can be. For a great many trained networks, there exist input perturbations small enough to be invisible on a display~\cite{moosavi2016deepfool}, inaudible through a loudspeaker~\cite{carlini2018audio}, or generally to be undetectable
, that can still change the model's output, for example, by
flipping the predicted class of a classifier, displacing or collapsing the bounding boxes of an object detector, or inducing failure in anomaly detectors.
\medskip

A signal processing reader is well placed to see why adversarial examples matter, and why they should not be filed away as an exotic curiosity of machine learning (and security) researchers. 
Adversarial examples are the worst-case version of such noisy inputs, and instead of showing \emph{how the system behaves under common noise}, they raise questions about \emph{what is the subtlest disturbance that breaks it}. Adversarial perturbations are relatively cheap to compute, yet rarely encountered by chance, unless their magnitude is large enough (see~\cite{fawzi2016robustness}). The gap between the typical and the worst-case perturbation justifies the adoption of the terminology \emph{attack}, which we embrace for its economy, while observing that nothing in what follows requires intent.
\medskip

A deployed safety-critical system is therefore only \emph{as reliable as the measurement of its robustness}, defined as the ability to tolerate input perturbations. Testing against a battery of known cases is not a reliable measure, since it characterizes robustness against known scenarios; however, it does not provide information nor guarantees about perturbations that are not tested. Similarly, methods that \emph{robustify} a model (named \emph{defenses}, for consistency with the security terminology) inherit the same limitation whenever they offer no guarantee on the resulting robustness, and the history of the field is populated with methods that seem to claim robustness but fail under more rigorous evaluations, namely improving the attack search algorithms. Defenses that also provide guarantees on the model's robustness tackle this weak point directly: they offer \emph{certifiable robustness}, a guarantee, computable on the deployed model, that no perturbation below a stated magnitude can change the outcome outside a tolerance.
\medskip

\noindent\textbf{This article} is a tutorial on \emph{Lipschitz-bounded} neural networks, the family of models that deliver such a guarantee by design at the lowest price currently known, i.e., a single forward pass, and without external algorithms. Broader surveys of certified robustness already exist, from the state of knowledge of Li~et~al.~\cite{li2023sok}, which spans verification, randomized smoothing and Lipschitz-based methods, to the review of Z\"uhlke~and~Kudenko~\cite{zuhlke2025adversarial}, which looks at robustness through a Lipschitz lens but is mostly concerned with estimating the constant of an unconstrained model. This tutorial instead focuses on Lipschitz-bounded models, deepening algorithmic and theoretical detail a practitioner needs to build, train, and certify one.
\medskip


The reason such a guarantee comes at this price is a nineteenth-century inequality: a bound on how much a transformation can amplify an input perturbation---the worst-case gain, i.e., its Lipschitz constant---directly limits how far the input can vary before the output changes more than a certain tolerance. In classification, this bound is determined by the margin between the winning score and its closest competitor, which is already available from a single forward pass and therefore turns this margin into a guaranteed perturbation radius: below that radius, the prediction cannot change. The problem thus reduces to two tasks: (i) building networks with a known Lipschitz constant, and (ii) training them to produce large margins while still achieving good performance on the tasks. 
\section*{History and Background}
The systematic study of these worst-case perturbations began as a security question. Biggio~et~al.~\cite{biggio2013evasion} showed that classifiers based on support vector machines (SVM) or also neural networks can be evaded at test time by including a malicious perturbation to the input, framing it as the solution of a minimization problem. Independently, Szegedy~et~al.~\cite{szegedy2014intriguing} observed the same fragility in deep image classifiers and named such a perturbed input an \emph{adversarial example}: one that is perceptually indistinguishable from a legitimate one, on which the model errs. Furthermore, they provided a preliminary explanation, reading the network as a cascade of operators, bounding their capability of propagating input perturbations (i.e., their \emph{Lipschitz constant}), and finding these bounds to be particularly large. This explains why two images that differ imperceptibly can therefore be associated with very different scores.

The decade that followed saw increasingly effective attacks for searching the neighborhood of an input with a perturbation that would change the model's outcome. An important step toward this direction is given by research on \emph{minimal adversarial perturbation}~\cite{carlini2017towards, pintor2021fast, brau2022minimal}, the closest input at which the decision flips, which measures directly how far a given sample lies from the decision boundary, and thus how much margin the model actually has, although approximated.

In parallel with the development of other attacking algorithms and discoveries of threats, the research community focused on methods to increase the reliability of the system under perturbation, i.e., using a security vocabulary, \emph{defenses}. 
In this regard, the canonical representative is \emph{adversarial training}~\cite{madry2018towards}, which descends directly from robust optimization methods and is written as a min--max problem. Such a formulation aims to minimize, over the model parameters, the worst loss attainable inside a perturbation ball around each training sample. The inner maximization can only be approximated, so what is actually minimized is a lower bound on the true worst-case loss. Adversarial training can not provide guarantees on the achieved robustness, which can only be measured on a test set exploiting known attacks. Other empirical defenses, such as input transformations, denoising front ends, and detectors for anomalous inputs, have followed a common pattern: a defense is proposed, shown effective against known attacks, and subsequently weakened by stronger or merely better-tuned ones. 
\begin{table}[t!]
    \centering
    \caption{Comparison of robustness-related strategies (i.e., defenses). Adversarial training increases robustness without providing guarantees, while Randomized smoothing certifies only in probabilistic terms. The \textbf{Employment} column indicates whether the method is applied at \textit{test} time, requires \textit{training} or both. }
    \label{tab:robustness_poles}
    \begin{tabular}{l|c|c|c}
        \toprule
        \textbf{Strategy}
        & \textbf{Robustify}
        & \textbf{Certify}
        & \textbf{Employment Time} \\
        \midrule

        Adversarial Training~\cite{madry2018towards}      & \cmark & \xmark & Train \\
        Formal Verification~\cite{katz2017reluplex}       & \xmark & \cmark & Test \\
        Randomized Smoothing~\cite{salman2019provably}    & \cmark & (\cmark) & Train/Test \\
        Lipschitz-Bounded Networks~\cite{anil2019sorting} & \cmark & \cmark & Train/Test \\

        \bottomrule
    \end{tabular}
\end{table}
\medskip

Instead, \emph{Certifiable} defenses return a guarantee together with the prediction.  Among these strategies, \emph{Verification methods}~\cite{katz2017reluplex,wong2018provable} encode the network, along with the largest admissible perturbation, as a system of constraints and hand the solution to an (mixed-integer) optimization solver~\cite{botoeva2020efficient}, whose cost grows with the number of neurons. However, complete verification, which is positive if and only if the prediction is in fact robust, is NP-complete for common ReLU feedforward DNNs~\cite{katz2017reluplex}, ruling out any (soft) real-time deployment on a network of interesting size. Recent verifiers therefore solve a relaxed problem, e.g., using a convex outer approximation of the reachable set~\cite{wong2018provable}, which is cheaper but not yet sufficiently fast for soft real-time systems. On the other hand, among certifiable methods, \emph{Randomized smoothing}~\cite{cohen2019certified,salman2019provably} takes a base classifier and produces a new one, named \emph{smoothed classifier}, which operates by adding Gaussian noise to the input and producing the classification by choosing the class with the majority vote. The smoothed classifier so defined is provably well-behaved, as it yields a certified input tolerance that grows with the confidence of the vote. The construction applies to any base model as a black box. However, it also carries two prices that strongly hinder its application: (i) the smoothed classifier is not an explicit function and requires a high computational cost to be processed, and (ii) the guarantees are probabilistic, where larger confidence requires a higher cost. Finally, a third family of defenses with guarantees is given by Lipschitz-bounded networks, which overcome the aforementioned limitations, and are described in detail in the next section. \Cref{tab:robustness_poles} provides an overview of the aforementioned methods. 

\subsection*{Robustness and Certification}
\label{sec:intro:decoupling}
Robustness, certification, and verification are sometimes not used consistently across communities. The two notions below and the definitions that follow provide the means to go through this tutorial.
\medskip

\noindent\textbf{Robustness} (Informal).
\textit{A model is robust at an input if every sufficiently small perturbation of that input leaves its outcome unchanged, or perturbed within a certain tolerance.}\medskip

\noindent\textbf{Certifiability} (Informal).
\textit{It is the ability to establish, by computation on the deployed model, up to a prescribed confidence level, that robustness holds at a given input.}
\medskip

It is worth belaboring the point that these two are logically independent. A model trained with aggressive augmentation may be genuinely hard to deceive and yet offer no mechanism whatsoever for proving it: robust but not certifiable. Conversely, a linear classifier is trivially certifiable, as the distance of an input from the decision boundary has a closed-form expression, but if its margins are small, the certificates it hands out are tiny, and the model is certifiable but not robust. 
Since both robustness and certification require committing to what ``small'' means, and hence to what metric spaces the input data lie in, henceforth, we consider a general metric space $\mathcal{X}$ equipped with a distance $d_\mathcal{X}$. 
\begin{defn}[Robustness of a classifier]
Given an input metric space $(\mathcal{X},d_\mathcal{X})$ and a positive noise level $\varepsilon$, we say that a classification model $f:\mathcal{X}\to \mathcal{Y}$, where $\mathcal{Y}$ is discrete, is $\varepsilon$-robust at a point $x\in \mathcal{X}$ if
\begin{equation}
    \forall x'\in\mathcal{X},\quad d_\mathcal{X}(x,x') \le \varepsilon \quad\Longrightarrow\quad f(x)=f(x').
    \label{eq:robust}
\end{equation}
An $\varepsilon$ such that $f$ is $\varepsilon$-robust at $x\in\mathcal{X}$ is said to be a \emph{radius} of $x$.
\end{defn}

This definition raises two natural questions. First, given a fixed magnitude $\varepsilon$, how hard is it to \emph{prove} that \Cref{eq:robust} holds? Second, given an input $x$, how hard is it to \emph{find} a noise-level $\varepsilon$ for which \Cref{eq:robust} holds? These two questions lead directly to the notions of \emph{verification} and \emph{certification}, which we state formally below.

\begin{defn}[Verification]
A \emph{verifier} $V$ is a program that takes as input a triple $(f,x,\varepsilon)$, within a classifier $f:\mathcal{X}\to\mathcal{Y}$, an input $x$, and a noise-level $\varepsilon$,  and establishes that $f$ is $\varepsilon$-robust at $x$, or returns the answer \emph{unknown}. A verifier is \emph{complete} if, whenever $f$ is not $\varepsilon$-robust at $x$, it provides a counterexample, i.e., a point $x'\in\mathcal{X}$ with $d_\mathcal{X}(x,x')\le\varepsilon$ such that $f(x')\ne f(x)$, and \emph{incomplete} otherwise.
\end{defn}

Verification is particularly hard to solve in a strict sense: Katz~et~al.~\cite{katz2017reluplex} proved, using instruments of \emph{satisfiability modulo theories}, that complete verification of ReLU-based neural networks is equivalent to the verification of the satisfiability of logical propositions, and hence is NP-complete with respect to the input dimension and the number of parameters of the model. For this reason, current literature on verification is focused on incomplete verification~\cite{wong2018provable}.

\begin{defn}[Certification]
A classification model $f:\mathcal{X}\to\mathcal{Y}$ is \emph{certifiable} if there exists a program that, given an input $x\in\mathcal{X}$, returns a (non zero) robustness radius of $x$, $C_f(x)$, which we name \emph{certificate}.
\end{defn}

Certification and verification are two views of the same problem. Unlike verification, certification only has to return one radius at a sample for which the classification is robust by definition, without the need of running an external verification algorithm to assess its robustness. 
Note that a complete verifier could be theoretically exploited to yield a certificate by bisection on $\varepsilon$ over a bounded input space, where each call of the verifier halves the upper bound of the robustness radius.\medskip


Finally, we note that the definition of certification we gave establishes that a prediction is \emph{stable}, not that it is \emph{correct}. If the model is confidently wrong at $x$, a certificate guarantees only that it remains confidently wrong throughout a neighborhood of $x$. Certifying correctness would require a notion of ground truth available to the model at inference time, and no current paradigm comes close. 
To obtain a single figure of merit, we fix a radius $\varepsilon$ and measure the probability that a sample is both correctly classified and robust; this is the metric used throughout the literature and throughout this article.

\begin{defn}[Robust Accuracy]
    Let $\mathcal{D}$ be a distribution over $\mathcal{X}\times\mathcal{Y}$     and $f:\mathcal{X}\to\mathcal{Y}$ a classification model. The     $\varepsilon$-\emph{robust accuracy} of $f$ is
    \begin{equation}
        \mathrm{Acc}(f,\varepsilon) \coloneqq
        \mathbb{P}_{(X,Y)\sim\mathcal{D}}
        \bigl\{f(X)=Y \,\land\, f \text{ is } \varepsilon\text{-robust at } X\bigr\}.
    \end{equation}
    Since every classifier is $0$-robust at every point,     $\mathrm{Acc}(f,0)$ coincides with the standard accuracy of $f$.
\end{defn}

In practice, $\mathrm{Acc}$ may not be operatively computable. Indeed, verifying $\varepsilon$-robustness in a complete sense is tough and depends on a verifier, and $\mathcal{D}$ is only accessible through a finite test set. An estimation of this metric is given by  \emph{certified robust accuracy} for certifiable models, which is a lower bound of the robust accuracy and the fraction of test samples that are correctly classified and for which $\varepsilon$ is a certificate. We call it $\mathrm{CRA}(f,\varepsilon)$.
\medskip

\subsection*{Certification via Lipschitz Bound}

The Lipschitz condition bounds the variation in the output in terms of the variation of the input. 
\begin{defn}[Lipschitz condition]
    Let $(\mathcal{X},d_\mathcal{X})$ and $(\mathcal{Y},d_\mathcal{Y})$ be two metric spaces. A map $f:\mathcal{X}\to\mathcal{Y}$ is \emph{Lipschitz} if there exist a constant $L\in\mathbb{R}$, such that
    \begin{equation}
        d_\mathcal{Y}(f(x),f(x'))\le L\,d_\mathcal{X}(x,x'),\quad \forall x,x'\in\mathcal{X},
    \end{equation}
    and in this case we say that $f$ is $L$-lipschitz.
\end{defn}


A classification model $f$, however, is more than a map between two metric spaces, and it is based on a continuous \emph{score function} $F$ that maps $\mathcal{X}$ onto an $\ell_p$-metric space $(\mathbb{R}^C,\ell_p)$, assigning a score to each of the $C$ classes, where the predicted class coincides with the index of the largest component. As the input crosses a decision boundary, the winning index changes abruptly; this makes $f$ discontinuous. Since every Lipschitz map is continuous, no (non-constant) classifier can be Lipschitz in the strict sense. For this reason, whenever we say for brevity that a classifier is Lipschitz, we mean that its score function $F$ is Lipschitz. From this, we formalise below  the main result of the tutorial, which  can be informally written as: 
\vspace{1em}

\noindent\emph{``A classifier with a known Lipschitz constant is (globally) certifiable at the cost of one model's forward''}.

\begin{thm}[Certificate of Lipschitz Classifiers]
    \label{thm:certification}
    Let $F:\mathbb{R}^d\to\mathbb{R}^C$ a score function between $\ell_p$- spaces with known lipschitz constant $L$, and let $f(x)=\argmax_i F_i(x)$ the induced classifier, then
    \begin{equation}
        \forall x\in\mathbb{R}^d,\quad C_f(x)\coloneqq \frac{F_y(x) - \max_{j\ne y} F_j (x)}{L\,\sqrt[q]{2}}\quad\text{is a certificate of $x$, }
        \label{eq:certificate}
    \end{equation}
    where $F_y(x)$ is the largest output component, and $q$ is the coniugate number of $p$, i.e., $\frac1p+\frac1q=1$. In other terms, applying the definition of certificate, for each $\delta$ such as $\|\delta\|_p\le C_f(x)$, then $f(x+\delta)=f(x)$.
\end{thm}

A formal proof of this statement is given in \cite[Theorem. 7]{li2019preventing}. Intuitively, the difference between the two largest output components is a direct measure of how much the score can vary before the predicted class changes; since this variation is proportional, through $L$, to a variation in the input space, rescaling it by $L$ yields the certificate. An alternative demonstration could be deduced by observing that, for a differentiable function $F$, an upper bound of the Lipschitz constant of the difference $F_i-F_j$ is given by the largest gradient $L^*_i\coloneqq \max_{x\in\mathcal{X}} \|\nabla (F_i-F_j)(x)\|_q$, which is the denominator of the Hein–Andriushchenko result on robustness~\cite[Theorem 1]{hein2017formal}.


On its own, the certificate in \eqref{eq:certificate} is of little practical use. The reason is that being Lipschitz is not a rare property: every feed-forward neural network built from standard layers (attention layers aside, a point we return to in the next section) is Lipschitz. However, after training by gradient descent, the network comes with some constant $L$, but its value is unknown. While a simple and cheap estimate of $L$ is obtained by multiplying the Lipschitz constants of the individual layers, this typically results in an overly pessimistic estimate, i.e., a large $L$, which in turn produces a small certificate $C_f$. Having a better estimate, however, does not automatically increase the certificate magnitude, as such a desired property has not been accounted for in the training process that should instead actively push the boundary far from the input samples.
Therefore, Theorem~\ref{thm:certification} alone is not sufficient, and what a recipe for certifiable robustness needs is: (i) a method to produce models with a known Lipschitz constant, and (ii) one to make its certificates large enough to be practically useful. This is the aim of the following sections.
\begin{figure}[!h]
\centering
\fbox{\begin{minipage}{0.94\linewidth}
\smallskip
\textbf{Who was Lipschitz?}
\smallskip

\par Rudolf Otto Sigismund Lipschitz (K\"onigsberg 1832 -- Bonn 1903) took his doctorate in Berlin under the supervision of Dirichlet and spent most of his career as professor of mathematics at Bonn, working across analysis, number theory, and Clifford algebras. 
The condition that carries his name came out as a solution of an intriguing problem: Under what condition does an ordinary differential equation $\dot x=f(x)$ have a unique solution? Continuity of $f$ turns out not to be enough, and Lipschitz's answer was to demand that
the output of $f$ not change more than linearly faster than its input. Condition that survives in every textbook as the hypothesis of the
Picard--Lindel\" of theorem. The inequality actually appeared a decade before the differential-equations work. Lipschitz first wrote it down in his 1864 study of the convergence of Fourier series (\emph{De explicatione per series trigonometricas}, Crelle's Journal), in the form $|f(x)-f(y)| \le M\,|x-y|^\alpha$ (which we would now call a H\"older condition), where the Lipschitz condition is the particular case with  $\alpha=1$.
\smallskip
\end{minipage}}
\end{figure}

\subsection*{Does the certificate approximate the boundary distance?}
From a geometrical perspective, the certificate discussed so far, $C_f(x)$, can be interpreted as a lower bound on the distance of $x$ from the decision boundary.
Let $\mathcal{B}$ be the decision boundary and denote by $\mathrm{d}_\mathcal{B}(x)$ the Euclidean distance of $x$ from $\mathcal{B}$. The distance function $\mathrm{d}_\mathcal{B}$ satisfies two properties. First, by construction, $\mathrm{d}_\mathcal{B}(x)$ is the \emph{best} (i.e., largest) possible certificate at $x$. Second, it satisfies a particularly strict global property: the gradient is not simply bounded; it is exactly unitary almost everywhere, $\|\nabla \mathrm{d}_\mathcal{B}(x)\|_2=1$.
Now, in the binary case, write the model outcome as a single score $F=F_1-F_2$, which vanishes on the boundary, i.e., $\mathcal{B}=\{x\,:\,F(x)=0\}$. If $F$ is $1$-Lipschitz, then $|F(x)|\le \mathrm{d}_\mathcal{B}(x)$, and $F$ only satisfies $\|\nabla F\|_2\le 1$. Wherever the inequality is strict, $F$ grows more slowly than the distance, so the certificate underestimates it.
It's then natural to ask whether classifiers whose score coincides with the distance from the decision boundary exist. 
When $F$ is smooth with $\|\nabla F\|_2=1$ on an open set containing the boundary,
$F$ coincides with the signed distance from $\mathcal{B}$ near it~\cite{brau2023robust}. 
Classifiers designed to satisfy this property are named Signed Distance Classifiers~\cite{brau2023robust, bethune2023robust}, and for them the certificate equals the norm of the minimal adversarial perturbation.
\section*{Lipschitz-Bounded Networks}
\label{sec:lipnets}
\begin{figure*}[h!]
  \centering
  \includegraphics[width=\textwidth, clip, trim=0 2cm 0 0]{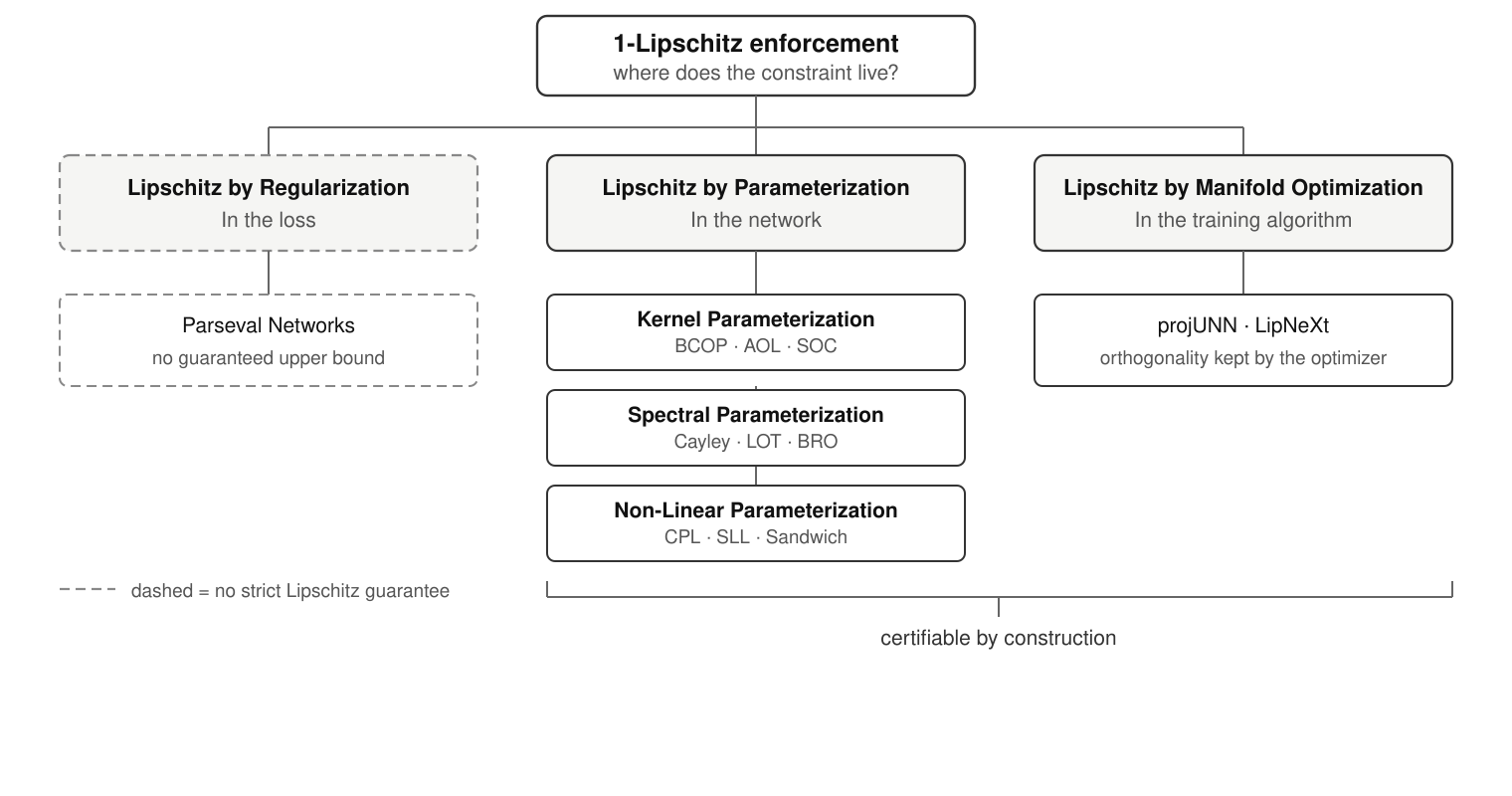}
  \caption{
  Taxonomy of $1$-Lipschitz layer constructions, organized by enforcement of the Lipschitz constraints.
  } 
  \label{fig:taxonomy}
\end{figure*}

How to build a neural model $F$ whose Lipschitz constant is known in advance? The main intuition comes from the composition property of Lipschitz functions. Precisely, if $F$ is obtained as a composition of Lipschitz functions, $F = f_K \circ \cdots \circ f_1$, then not only is $F$ Lipschitz, but also 
\begin{equation}
  \mathrm{Lip}(F) \;\le\; \prod_{k=1}^{K} \mathrm{Lip}(f_k),
  \label{eq:product}
\end{equation}
where $\mathrm{Lip}(f)$ denotes the smallest Lipschitz constant of $f$. \Cref{eq:product} is therefore a bound on the Lipschitz constant of the whole model, controlled independently by each layer's constant. An immediate consequence is that composing $1$-Lipschitz layers produces a $1$-Lipschitz neural network~\cite{prach2024lipschitz}, which is by far the most common design choice in the literature.
Affine functions and non-linear activations are the building blocks of feed-forward models. A linear layer $f(x) = Wx+b$ is Lipschitz with constant equal to the spectral norm of the weight matrix $\|W\|$, defined as its largest singular value, and the majority of activation functions, including ReLU, are 1-Lipschitz.
Although simple-looking, the literature on how to build these networks is large and fast-moving. We organize it around one question: \emph{where, when, and how is the Lipschitz constraint imposed?} See \Cref{fig:taxonomy} for an overview of the three main methodologies.
\medskip

We deliberately refrain from ranking these constructions by performance, since the most suitable choice depends on the scenario at hand, e.g., the available memory and compute budget, the architecture, and the target certification radius. We refer the interested reader to Prach et al.~\cite{prach2024lipschitz} for a systematic comparison of their memory footprint, speed, and certified robust accuracy.
\medskip

Before introducing the methodologies, we fix some notation. We denote by $\mathcal{O}(d)\coloneqq\{Q\in\mathbb{R}^{d\times d}\,:\,Q^{\top}Q=I\}$ the family of orthogonal matrices, and by
$\mathcal{L}(d)\coloneqq\{A\in\mathbb{R}^{d\times d}\,:\,\|A\|_2\le 1\}$ the set of matrices whose singular values are at most $1$, dropping the dimensionality $d$ whenever the
dimension is clear from the context. Since $\|A\|_2$ is the largest singular value of $A$, the elements of $\mathcal{L}(d)$ are exactly the linear maps that are $1$-Lipschitz, where the orthogonal represent a particular case, i.e., $\mathcal{O}(d)\subseteq\mathcal{L}(d)$.

\subsection*{Lipschitz by Regularization}
\label{sec:lipnets:regularization}
Cisse~et~al.~\cite{cisse2017parseval} introduced Parseval Networks, based on the intuition that affine layers with an orthogonal weight matrix, having all singular values equal to $1$, are 1-Lipschitz. To this end, they introduced a regularization term to each of the learnable weights of the network
\begin{equation}
    R_\beta(W)\coloneqq \frac{\beta}{2}\|W^\top W - I\|_F^2,
\end{equation}
where $\|\cdot\|_F$ represents the Frobenius norm. Optimizing such a term up to zero, whose convergence is provable (See the Sidebar on Regularization and Convergence), would produce exact orthogonality. The method has been extended to convolutions, noticing that they can be represented with matrices with a double-t\"oplitz structure, at the cost of losing the convergence property of the regularization term~\cite{wang2020orthogonal}. Nevertheless, an exact solution is still too expensive, as it should be repeated after each weight update; for this reason, only a few steps are computationally sustainable, making the result only close to orthogonality, losing a strict constraint on the Lipschitz constant. 
\begin{figure}[t!]
\centering
\label{side:convergence}
\fbox{\begin{minipage}{0.94\linewidth}
\smallskip
\textbf{Regularization and Convergence.}\\
Plain gradient descent on $R_\beta$ with
learning rate $\alpha=\beta=\frac{1}{2}$ produces the iteration
\[
W_{k+1} \;=\; W_k - \alpha\nabla R_\beta(W_k)
        \;=\; W_k\Bigl(I + \tfrac12\bigl(I - W_k^\top W_k\bigr)\Bigr),
\]
which is exactly the first-order ($p=1$) Bj\"orck--Bowie
orthogonalization~\cite{bjorck1971iterative}. 
The $R_\beta$ term has zero gradient at the fixed point $W^TW=I$, and singular values converge quadratically to $1$ provided $0<\sigma_i(W_0)<1$; in practice convergence is guaranteed by rescaling $W_0$ so that $\|W_0\|_2 \le 1$. In other words, regularization with the right step is not an approximation of the projection onto the orthogonal matrices, but it coincides with the projection algorithm, truncated after a few steps.
\smallskip
\end{minipage}}
\end{figure}

\begin{figure}[h!]
  \centering
  \begin{tikzpicture}
    \drawblob
    \coordinate (wi)   at (-2.30,2.55);
    \coordinate (wt)   at (0.30,2.95);
    \coordinate (wn)   at (0.75,1.95);
    \coordinate (foot) at (0.72,1.34);
    \draw[gstep] (wi) -- (wt);
    \draw[pull]  (wt) -- (wn);
    \draw[aux]   (wn) -- (foot);
    \node[wpt] at (wi) {};  \node[wpt] at (wt) {};  \node[wpt] at (wn) {};
    \node[lbl, left=3pt of wi]  {$W_i$};
    \node[lbl, above=3pt of wt] {$\tilde W_{i+1}$};
    \node[lbl, left=3pt of wn]  {$W_{i+1}$};
    \node[slbl, below=2pt] at ($(wi)!0.5!(wt)$) {$-\alpha\nabla\ell(W_i)$};
    \node[slbl, right=3pt, red!65!black] at ($(wt)!0.5!(wn)$)
          {$-\alpha\nabla R_\beta(\tilde W_{i+1})$};
    \draw[aux] ($(wn)!0.5!(foot)$) -- (3.05,2.05);
    \node[slbl, gray!55!black, right=1pt] at (3.05,2.05)
          {$\mathrm{dist}(W_{i+1},\mathcal{O}(d))>0$};
  \end{tikzpicture}
  \caption{
  The weight $W_i$ lives in the ambient space, off $\mathcal{O}(d)$, the set of admissible orthogonal weight matrices
  A plain gradient step on the classification loss  $\ell$ takes it to $\tilde W_{i+1}$, in a direction that has no reason to respect the constraint;  the penalty term $R_\beta$ then pulls the iterate towards $\mathcal{O}(d)$, producing the updated weight $W_{i+1}$. Because the two components compete through the weight $\beta$, the iterate approaches the manifold but does not guarantee to reach it.}
  \label{fig:mech_reg}
\end{figure}
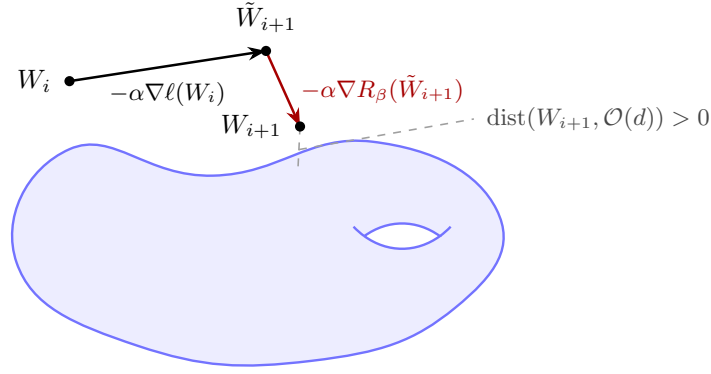

\Cref{fig:mech_reg} provides an intuition on regularization-based Lipschitz enforcement. The weights of the model are updated with a regular gradient-based method and regularized to remain close to the manifold of orthogonal matrices, represented by the blue blob. 

\subsection*{Lipschitz by Parameterization}
\label{sec:lipnets:parameterization}

Whereas regularization requires the optimizer to search for a 1-Lipschitz network, parameterization restricts the search space, excluding every network with an unbounded Lipschitz constant. In a parameterized $1$-Lipschitz layer, the weights applied to the input are not learned directly; instead, they are mapped from (free) parameters under a fixed map $\Phi$, depending on the method, that makes the layer $1$-Lipschitz for every choice of parameters, e.g., mapping the orthogonal space $\mathcal{O}(d)$.

Dense linear layers are particularly suited to this. For a given space dimension $d$, let $\Phi:\mathbb{R}^{d\times d}\to \mathcal{L}(d)$ be a map onto the constrained weights living in $\mathcal{L}(d)$, having singular values at most $1$. By construction, every dense layer of the form
\begin{equation}\label{eq:param}
    f_i\,:\,x\quad\mapsto\quad\Phi(\theta)x+b,\quad\forall\theta\in\mathbb{R}^{d\times d},\,b\in\mathbb{R}^d,
\end{equation}
is $1$-Lipschitz, while the learnable parameter $\theta$ remains unconstrained. The advantages are clear: gradient descent runs on $\theta$ without modification, so the layer can be dropped into any architecture without touching the training pipeline; moreover, at test time the parameterized weight $W=\Phi(\theta)$ can be computed once and cached, so that the cost of evaluating $\Phi$ disappears and inference is as cheap as for an unconstrained network. The second column of \Cref{tab:param_maps} collects the maps $\Phi$ proposed in the literature; for dense layers the methods differ only in the map they choose. The sidebar on \emph{exponential orthogonalization} provides a detailed example on an iconic case, the exponentiation of skew-symmetric matrices, which is also a link to the following method for lipschitz enforcement by manifold optimization.
\medskip

The differences appear when the map is extended to convolutions, which can be done in two ways, \emph{Kernel} and \emph{Spectral} Parameterizations, depending on what domain the parameterization operates.

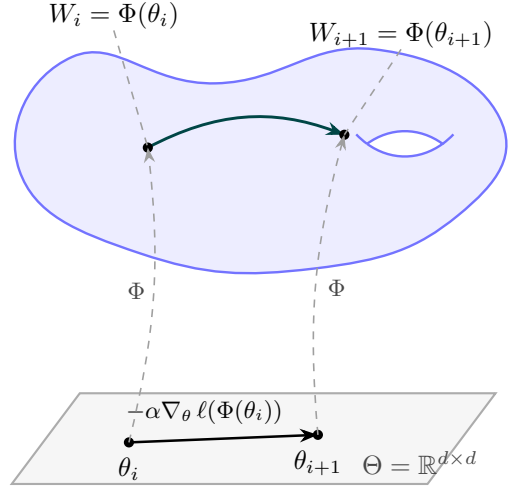
\begin{figure}[h!]
  \centering
  \begin{tikzpicture}
    \drawblob
    \coordinate (wi) at (-1.30,0.45);
    \coordinate (wn) at (1.30,0.62);
    \draw[geo] (wi) to[bend left=24] (wn);
    \node[wpt] at (wi) {};  \node[wpt] at (wn) {};
    \node[lbl] at (-1.75,2.20) {$W_i=\Phi(\theta_i)$};
    \node[lbl] at (2.05,1.95)  {$W_{i+1}=\Phi(\theta_{i+1})$};
    \draw[aux] (-1.75,2.00) -- (wi);
    \draw[aux] (2.05,1.75) -- (wn);
    \begin{scope}[yshift=-3.4cm]
      \draw[fill=gray!7, draw=gray!65, line width=0.8pt]
        (-3.1,-0.60) -- (2.4,-0.60) -- (3.3,0.60) -- (-2.2,0.60) -- cycle;
      \node[gray!55!black, lbl] at (2.30,-0.32) {$\Theta=\mathbb{R}^{d\times d}$};
      \coordinate (ti) at (-1.55,-0.05);
      \coordinate (tn) at (0.95,0.05);
      \node[wpt] at (ti) {};  \node[wpt] at (tn) {};
      \draw[gstep] (ti) -- (tn);
      \node[lbl, below=2pt of ti] {$\theta_i$};
      \node[lbl, below=2pt of tn] {$\theta_{i+1}$};
      \node[slbl, above=2pt] at ($(ti)!0.40!(tn)$)
        {$-\alpha\nabla_\theta\,\ell(\Phi(\theta_i))$};
    \end{scope}
    \draw[aux, -{Stealth[length=2mm]}] (ti) to[bend right=10] (wi);
    \draw[aux, -{Stealth[length=2mm]}] (tn) to[bend left=10] (wn);
    \node[slbl, gray!55!black] at (-1.45,-1.45) {$\Phi$};
    \node[slbl, gray!55!black] at (1.2,-1.40) {$\Phi$};
  \end{tikzpicture}
  \caption{Weights $W_i$ lives in the manifold $\mathcal{L}$, while the learnable parameters $\theta_i$ lives in the flat unconstrained parameter space $\Theta$. The update of the parameter happens through a gradient step that
  moves $\theta_i$ to $\theta_{i+1}$. The map $\Phi$ then sends each parameter to a weight matrix that lies in $\mathcal{O}(d)$ by construction, so both the updated iterate $W_{i+1}=\Phi(\theta_{i+1})$ is admissible.}
  \label{fig:mech_param}
\end{figure}

\paragraph*{Kernel Parameterization} The mapping $\Phi$ takes the free kernel parameter $\theta\in\mathbb{R}^{c\times c\times k\times k}$ and returns a weight $k\times k$ kernel, which is then applied with a standard convolution. This preserves some advantages of the dense case, including the cached inference cost. Miyato~et~al.~\cite{miyato2018spectral} proposed the first and more intuitive approach, where the kernel's singular values are normalized through the power method, used to estimate $\|W\|_2$, i.e., the largest singular value of the linear map $x\mapsto W\star x$, where $W\in\mathbb{R}^{c\times c\times k\times k}$, hence producing $\Phi(W)=W/\|W\|_2$. Singular values of a convolutional layer can also be computed exactly via SVD decomposition; see~\cite{sedghi2018singular}, although typically not without sacrificing computational cost during training. Furthermore, this category includes AOL~\cite{prach2022almost}, BCOP~\cite{li2019preventing} and SOC~\cite{singla2021skew}; see the third column of \Cref{tab:param_maps} for the corresponding maps. SOC is a partial exception and does not allow caching; indeed, the exponentiated convolution is approximated by a truncated series and computed online on each input, so the weight kernel cannot be stored, and the inference cost grows with the truncation order.

\paragraph*{Spectral Parameterization} Each mapping $\Phi$ valid for dense layers can be applied to convolutions by moving to the frequency space. Indeed, by the convolution theorem, a circular convolution becomes a multiplication under the DFT, $\mathcal{F}(\theta\star x)=\mathcal{F}\theta\cdot\mathcal{F}x$, 
where, for $d$ channels, the product is taken independently for each frequency, where, at each frequency $\omega$, the kernel in the frequency space is a $d\times d$ complex matrix $\widehat{\theta}_\omega$ acting on the $d$-dimensional vector $\widehat{x}_\omega$. Since the DFT is orthogonal (i.e., represented by a unitary matrix), the convolution is $1$-Lipschitz if and only if every $\widehat{\theta}_\omega$ has spectral norm at most one. A parameterization $\Phi$ for matrices can therefore be applied in the frequency domain to each block $\widehat{\theta}_\omega$ and brought back to the input domain, obtaining $\mathcal{F}^{-1}\big\{\Phi(\widehat{\theta}_\omega)\big\}_{\omega}$ that is 1-Lipschitz. \Cref{tab:param_maps} includes several methods known in the literature.

However, in practice, mapping back the constrained blocks into the kernel space would produce kernels with full spatial support (a kernel size that matches the size of the input image) which is computationally prohibitive. For this reason, the multiplication is directly performed in frequency space and mapped back to the input, as follows:
\begin{equation}\label{eq:param_freq}
    f_i:x\;\mapsto\;\mathcal{F}^{-1}\big\{\Phi(\widehat{\theta}_\omega)\,\widehat{x}_\omega\big\}_{\omega}+b ,
\end{equation}
which is $1$-Lipschitz for every unconstrained $\theta$. The cost is one forward and one inverse DFT per layer, also at inference, while the cost of $\Phi$ could be limited at test time by caching of $\Phi(\hat\theta_\omega)$.

\begin{figure}[!t]
\centering
\fbox{\begin{minipage}{0.94\linewidth}
\smallskip
\textbf{The exponential orthogonalization $\Phi(\theta)=\exp(\theta-\theta^{\top})$}\\
\label{side:exp}
\smallskip
The matrix exponential of a skew-symmetric matrix $A$, i.e., such that $A=-A^{\top}$, is
orthogonal. The statement follows directly from the definition of the exponential
(see, e.g.,~\cite{hall2015lie}) and applies to $A=\theta-\theta^{\top}$, which is skew for every $\theta$.
Through the lens of differential geometry, the family of orthogonal matrices
$\mathcal{O}(d)$ form a manifold in
$\mathbb{R}^{d\times d}$, where the space of skew-symmetric matrices $\mathfrak{so}(d)$ constitute its tangent space at the identity, $\mathfrak{so}(d)=T_{I}\,\mathcal{O}(d)$.
\par In this space, the curve $t\mapsto\exp(tA)$, starting at $I$ with velocity $A$, is known as the \emph{geodesic} in direction $A$, i.e., the straightest and (locally) shortest possible path living in the surface. Hence, $\Phi(\theta)$ not only maps the parameter into the space of $1$-Lipschitz matrices, but furthermore into the manifold of (special) orthogonal matrices, as the endpoint of the geodesic from $I$ among $\theta-\theta^\top$.
\par This principle underpins the third category of Lipschitz-bounded models, where Lipschitz enforcement is imposed in the training algorithm to move along the manifold during optimization.
\smallskip
\end{minipage}}
\end{figure}

\paragraph*{Non-Linear Parameterization} A few other parameterized methods, CPL~\cite{meunier2022dynamical}, SLL~\cite{araujo2023unified} and Sandwich~\cite{wang2023direct} falls outside both classes. In these methods, the parameterization involves the activation, which is built into the layer. These layers are exactly $1$-Lipschitz but are not affine transformations (See the third section of \Cref{tab:param_maps} for more details). Furthermore, CPL and SLL are characterized by involving a residual term, which has been empirically proved to improve performance in Prach~et~al.~\cite{prach2024lipschitz}, and constitutes a sufficient condition for density results (see \cite{murari2025approximation}), as discussed in the dedicated section.

\begin{table*}[b!]
\centering
\caption{The parameterization map $\Phi$ for dense ($\theta\in\mathbb{R}^{c\times c}$) and
convolutional ($\theta\in\mathbb{R}^{c\times c\times k\times k}$) parameters. For kernels,
$\theta^{\top}$ denotes the adjoint kernel and
$\mathcal{F}$ the spatial DFT, so that $\mathcal{F}\theta$ is a collection of
$c\times c$ complex blocks to which the dense map is applied blockwise
(with $^{\ast}$ in place of $^{\top}$). $\sigma$ denotes the ReLU, $b$ a bias.}
\label{tab:param_maps}
\small
\renewcommand{\arraystretch}{1.5}
\begin{tabular}{l|c|c}
\toprule
\textbf{Layer} & \textbf{Dense:} $\Phi(\theta)$ & \textbf{Convolution:} $\Phi(\theta)$ \\
\midrule
\multicolumn{3}{c}{\emph{Kernel Parameterization}}\\
\midrule
AOL~\cite{prach2022almost}
 & $\mathrm{AOL}(\theta)=\theta\,\Big(\underbrace{\textstyle\sum_{j}\big|\theta^{\top}\theta\big|_{ij}}_{D(\theta)}\Big)^{-1/2}$
 & $\mathrm{AOL}(\theta)=\theta\,\Big(\underbrace{\textstyle\sum_{j,\,p}\big|\theta^{\top}\!\star\theta\big|_{ij,p}}_{D(\theta)}\Big)^{-1/2}$ \\
BCOP~\cite{li2019preventing}
 & $BB(\theta)=A_k$, $A_{i+1}=A_i\big(I+\tfrac12(I-A_i^{\top}A_i)\big)$
 & $\mathrm{BCOP}(\theta)=[\mathrm{BB}(\theta_0),\dots,\mathrm{BB}(\theta_{2k-2})]$\textsuperscript{$\dagger$}\\
SOC~\cite{singla2021skew}
 & $\exp(A)=\sum_{j\ge 0}\frac{A^{j}}{j!}$,  $A=\theta-\theta^{\top}$
 & $\exp_{\star}(A)=\sum_{j\ge 0}\frac{A^{\star j}}{j!}$, $A=\theta-\theta^{\top}$ \\
Cholesky~\cite{hu2024recipe}
 & $\mathrm{Chol}(\theta)=\theta\,L^{-\top}$, $\theta^{\top}\theta=LL^{\top}$
 & --- \\
\midrule
\multicolumn{3}{c}{\emph{Spectral Parameterization}}\\
\midrule
Cayley~\cite{trockman2021orthogonalizing}
 & $\mathrm{Cay}(\theta)=(I-\theta+\theta^{\top})(I+\theta-\theta^{\top})^{-1}$
 & $\mathcal{F}^{-1}\big(\mathrm{Cay}(\mathcal{F}\theta)\big)$ \\
LOT~\cite{xu2022lot}
 & $\mathrm{Pol}(\theta)=\theta\,(\theta^{\top}\theta)^{-1/2}$
 & $\mathcal{F}^{-1}\big(\mathrm{Pol}(\mathcal{F}\theta)\big)$ \\
BRO~\cite{lai2025bronet}
 & $\mathrm{Ref}(\theta)=I-2\,\theta(\theta^{\top}\theta)^{-1}\theta^{\top}$
 & $\mathcal{F}^{-1}\big(\mathrm{Ref}(\mathcal{F}\theta)\big)$ \\
\midrule
\multicolumn{3}{c}{\emph{Non-Linear Parameterization (layer $f(x)$ reported instead of $\Phi$)}}\\
\midrule
CPL~\cite{meunier2022dynamical}
 & $\mathrm{CPL}(x)=x-\frac{2}{\|\theta\|_2^{2}}\,\theta^{\top}\sigma(\theta x+b)$
 & $\mathrm{CPL}(x)=x-\frac{2}{\|\theta\|_2^{2}}\,\theta^{\top}\!\star\sigma(\theta\star x+b)$ \\
SLL~\cite{araujo2023unified}
 & $\mathrm{SLL}(x)=x-2\,\theta D(\theta)^{-1}\sigma(\theta^{\top}x+b)$
 & $\mathrm{SLL}(x)=x-2\,\theta\star D(\theta)^{-1}\sigma(\theta^{\top}\!\star x+b)$ \\
SNW~\cite{wang2023direct}
 & $\mathrm{SNW}(x) = A^{\top}\sigma(Bx+b)$,\quad $[A\;B]=\mathrm{Cay}(\theta)\textsuperscript{$\ddagger$}$
 &  
 $\mathcal{F}^{-1}(\mathrm{SNW}(\hat x))$\textsuperscript{$\S$}\quad \\
\bottomrule
\end{tabular}
\vspace{3pt}
\begin{minipage}{0.97\textwidth}
\footnotesize
\vspace{5pt}
\textsuperscript{$\dagger$}Composition of $2k-1$ orthogonalized $1\times 1$ kernels.\\
\textsuperscript{$\ddagger$}Extended Cayley transform of~\cite{wang2023direct}.\\
\textsuperscript{$\S$}$SNW(\hat x)=\hat A^{\top}\mathcal{F}(\sigma(\mathcal{F}^{-1}(\hat B\,\hat x+b))$, where $\hat x=\mathcal{F}(x)$, and  $[\hat A\;\hat B]=\mathrm{Cay}(\mathcal{F}\theta)$\\
\end{minipage}
\end{table*}

One property is shared by every method in this subsection; the construction $\Phi$ sits {inside} the network. It is evaluated on every forward pass and differentiated through on every backward pass, for every layer, for the entire duration of training and, unless the weights are frozen and $\Phi(\theta)$ cached, at inference too. This is where the majority of the literature on certifiable robustness via Lipschitz-Bounded Networks lives. The \Cref{fig:mech_param} summarizes this idea.

\subsection*{Lipschitz by Manifold Optimization}
\label{sec:lipnets:manifold}

\begin{figure}[t]
  \centering
  \begin{tikzpicture}
    \drawblob
    \coordinate (wi) at (-0.55,0.35);
    \draw[tangent] ($(wi)+(-1.70,-0.10)$) -- ($(wi)+(0.80,-0.60)$)
                -- ($(wi)+(2.00,0.35)$) -- ($(wi)+(-0.50,0.85)$) -- cycle;
    \node[orange!75!black, slbl] at ($(wi)+(-1.15,0.60)$) {$T_{W_i}\mathcal{O}(d)$};
    \coordinate (gtip) at ($(wi)+(1.15,1.95)$);
    \coordinate (ptip) at ($(wi)+(1.42,0.12)$);
    \draw[gstep] (wi) -- (gtip);
    \draw[aux]   (gtip) -- (ptip);
    \draw[geo]   (wi) -- (ptip);
    \node[lbl, above=2pt of gtip] {$\nabla\ell(W_i)$};
    \node[slbl, teal!45!black, below right=-1pt and 0pt of ptip] {$\mathfrak{g}$};
    \coordinate (wn) at (-2.15,-0.42);
    \draw[geo, line width=1.3pt] (wi) to[bend right=16] (wn);
    \node[slbl, teal!45!black] at (-1.05,-0.75) {$-\alpha\mathfrak{g}$};
    \node[wpt] at (wi) {};  \node[wpt] at (wn) {};
    \node[lbl, above left=1pt and -1pt of wi] {$W_i$};
    \node[lbl] at (-2.80,-1.55) {$W_{i+1}=\exp_{W_i}(-\alpha\mathfrak{g})$};
    \draw[aux] (-2.55,-1.35) -- (wn);
  \end{tikzpicture}
  \caption{The weights $W_i$ starts on the manifold $\mathcal{O}(d)$. The update is performed not in the ambient space (i.e., using the Euclidean gradient $\nabla\ell(W_i)$), since it generally points off the manifold. However, the update is projected onto the tangent space $T_{w_i}\mathcal{O}(d)$ (shaded), giving the Riemannian gradient $\mathfrak{g}$, the steepest direction {along} the manifold. The step is then taken by the \emph{geodesic exponential map}, which travels along the shortest path on $\mathcal{O}(d)$ leaving $W_i$ with initial direction $-\alpha\mathfrak{g}$, tracing the curve to   $W_{i+1}=\exp_{W_i}(-\alpha\mathfrak{g})$; the iterate has moved and is still admissible.}
  \label{fig:mech_manifold}
\end{figure}
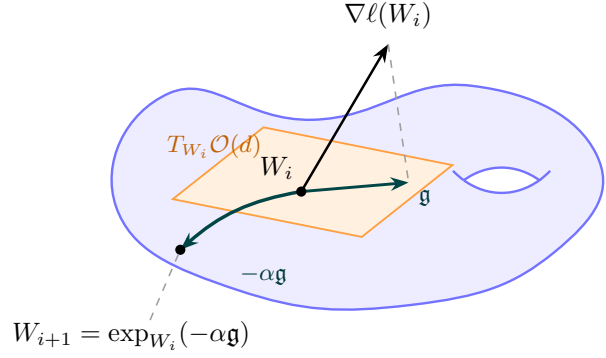

The two previous families enforce the Lipschitz condition either in the loss, through the penalty $R_\beta$, or in the network's parameters, through the map $\Phi$. This third family instead hands it to the optimizer itself: a plain feed-forward model $F=f_K\circ\cdots\circ f_1$ is initialized so that its weight matrices lie on the manifold $\mathcal{O}(d)$ of orthogonal matrices, and the update rule of the optimizer is designed so that orthogonality is preserved at every iteration. The mechanics are summarized in \Cref{fig:mech_manifold}.

For each weight $W\in\mathcal{O}(d)$, the Euclidean gradient $\nabla\ell(W)$ generally points off the manifold, so a classical step $W_{\mathrm{new}}\gets W-\alpha\nabla\ell(W)$ would bring $W_{\mathrm{new}}$ outside $\mathcal{O}(d)$, and the whole model $F$ would lose the $1$-Lipschitz property. A valid update is produced in two stages. First, the gradient is projected onto the tangent space to $\mathcal{O}(d)$ at $W$,
\begin{equation}
    T_W\,\mathcal{O}(d) = \{A\in\mathbb{R}^{d\times d}\,:\, W^\top A + A^\top W = 0\},
\end{equation}
which reduces to the skew-symmetric matrices $\mathfrak{so}(d)$ discussed in the sidebar when $W=I$. The projection of $\nabla \ell(W)$ onto $T_W\mathcal{O}(d)$ has a closed form $\mathfrak{g}\in T_W\mathcal{O}(d)$, known as \emph{Riemannian gradient}, defined by
\begin{equation}\label{eq:riem_grad}
    \mathfrak{g} \coloneqq \nabla\ell(W) - W\,\mathrm{sym}\big(W^\top\nabla\ell(W)\big),
    \qquad \mathrm{sym}(B)\coloneqq\tfrac12\big(B+B^\top\big),
\end{equation}
the steepest-descent direction that doesn't break the constraint. Second, since a simple addition along a tangent direction would also leave the manifold, the step is taken along $\mathcal{O}(d)$ rather than along a straight line, through a \emph{retraction}, i.e., a map from the tangent space back onto $\mathcal{O}(d)$. Several retractions are available, and they mirror the maps of \Cref{tab:param_maps}: the geodesic exponential map $W_{\mathrm{new}}=W\exp(-\alpha\,W^\top\mathfrak{g})$~\cite{hu2026lipnext}, the Cayley transform~\cite{trockman2021orthogonalizing}, and the polar factor $\tilde W(\tilde W^\top\tilde W)^{-1/2}$ of the Euclidean iterate $\tilde W=W-\alpha\nabla\ell(W)$, adopted by projUNN~\cite{kiani2022projunn}, which skips the tangent projection altogether and exploits a low-rank approximation of the gradient to keep small cost on each update. Taking the exponential map as a representative, since $W^\top\mathfrak{g}=\mathrm{skew}\big(W^\top\nabla\ell(W)\big)$ is skew-symmetric by construction, hence $\exp(-\alpha\,W^\top\mathfrak{g})$ is orthogonal and so is its product with $W$, the same property exploited by the SOC layer, now applied to the {update} rather than directly to the parameter. Whatever the retraction, the iterate never leaves $\mathcal{O}(d)$, so the layer is exactly $1$-Lipschitz at every step of training.

In conclusion, all the methods in this family operate on matrix-like weights only and not directly on kernels for convolutional layers: to the best of our knowledge, no retraction is available that acts directly on a $d\times d\times k\times k$ kernel while preserving the $1$-Lipschitz property of the resulting convolution. LipNeXt~\cite{hu2026lipnext} circumvents the issue by emulating a convolution through \emph{spatial mixing}, which utilizes a parameter-free \emph{shift} operator $\mathcal{S}$, and where each layer takes the form $x\mapsto R^\top\,\mathcal{S}(Rx)$, where $R$ is an orthogonal $c\times c$ matrix updated on the manifold as above. Since $\mathcal{S}$ is a permutation, hence orthogonal, and $R$ is kept orthogonal by the optimizer, the composition is exactly $1$-Lipschitz; the receptive field grows with depth, as in a stack of $3\times 3$ convolutions, but no kernel-like weight is learned. Whether a genuine manifold optimization for convolutional kernels can be deduced remains, to the best of our knowledge, an open question.

\subsection*{Attention Layers are not Lipschitz}
The attention mechanism has pushed the performance of language models and video language models, standing as an essential recipe of all the transformer-like architectures. Collecting a sequence of $n$ tokens in the rows of $x\in\mathbb{R}^{n\times d}$, a (single-head) attention layer is defined as
\begin{equation}\label{eq:attention}
\mathrm{Att}(x)=\mathrm{softmax}\!\left(\frac{x W_Q W_K^\top x^\top}{\sqrt{d}}\right) x W_V,
\end{equation}
where the scores $x_i^\top W_Q W_K^\top x_j$ represent {bilinear} function of the
input. Its presence makes the variation of the output with respect to an input $x$ unbounded, and in particular not Lipschitz. An informal argument for this can be obtained by observing that the bilinear form adjusts the weights $W_V$ to which the input is applied, so that the resulting weights are not fixed. More formally, one can consider the derivative of the layer and observe that the Jacobian is not bounded in any $\ell_p$ norm~\cite{kim2021lipschitz}.

Making attention Lipschitz requires avoiding the bilinear product at the core of the mechanism. The first attempts replace the dot product score with a negative squared distance, $-\|x_iW_Q-x_jW_K\|_2^2$, which is Lipschitz but whose bound degrades with sequence length~\cite{kim2021lipschitz}, or take the lighter route of rescaling the magnitude of the attention scores during training~\cite{dasoulas2021lipschitz}. At the architecture level, LipsFormer~\cite{qi2023lipsformer} substitutes a Lipschitz continuous surrogate for every transformer component: scaled cosine similarity for the dot product, centering normalization for layer normalization, whose gain diverges as the token variance vanishes, and weighted residual shortcuts. In conclusion, to the best of our knowledge, no transformer has been certified at radii comparable to the convolutional models, and their application is more suited to training stability rather than certification. Whether the Lipschitz property is an indispensable recipe for training capability is an open question.

\section*{Training Lipschitz-Bounded Networks}
\label{sec:training}

\begin{figure}[h!]
  \centering
  \includegraphics[width=\linewidth]{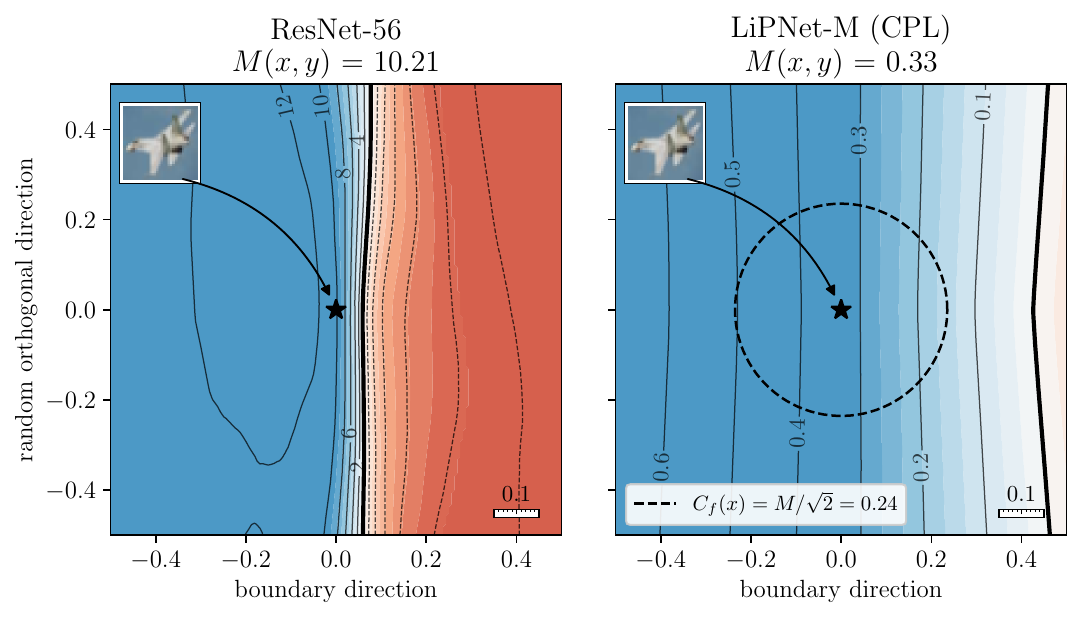}
  \caption{In scale representation of the margin landscape of a naive ResNet-56 and a Lipschitz-bounded classifier~\cite{prach2024lipschitz}  around a CIFAR-10 image on the 2D plane spanned by $-\nabla M(x,y)$ (the boundary direction) and a random orthogonal direction. Distances are measured in $\ell_2$ in pixel space, where the bold curve is the decision boundary $M=0$. For the ResNet-56, the margin drops from $10.2$ to $0$ within a surprisingly small (and unpredictable) distance from the boundary of $\approx0.07$, while for the 1-Lipschitz model it decreases slowly with a bounded rate, offering a certified radius $C_f(x)\approx 0.24$.}
  \label{fig:margin_landscape}
\end{figure}

The constructions of Lipschitz-bounded networks provide available certification by construction, at inference time. Nevertheless, still, a mechanism to obtain useful certificates is still needed, i.e., one that allows classification with guaranteed robustness for a sufficiently large radius. Reading the certification in \eqref{eq:certificate}, the certified radius of a sample is decided entirely by the numerator, the margin
\begin{equation}
  M(x,y) \coloneqq F_y(x) - \max_{j\neq y} F_j(x),
  \qquad
  C_f(x) = \frac{M(x,y)}{L\sqrt{2}}.
  \label{eq:margin}
\end{equation}
Figure~\ref{fig:margin_landscape} visualizes this quantity. For a 1-Lipschitz network, the margin can
decrease at most at rate $\sqrt{2}$ as the input moves, so the level curves of $M$ are spread apart and
every point within $C_f(x)$ keeps the prediction. A standard network may exhibit a much larger margin
at the sample and yet lose it within a far smaller distance, clarifying that without a known (and small) Lipschitz
constant, a large margin says nothing about robustness.

Hence, a $1$-Lipschitz network trained without accounting such a margin during training, may be certifiable but useless, as the certificates it emits are correct but close to zero.
Lipschitz-aware Loss function closes that gap between by-design and useful certification.

\subsection*{Lipschitz-Aware Loss Functions}
\label{sec:training:loss}

How could we induce a model to satisfy a high certifiable robust accuracy during training? The mainstream solution consists in considering an ad hoc loss function, which jointly makes the model achieve high accuracy and keeps the certification radius large. For Lipschitz-bounded models, this consists in keeping the margin between output components sufficiently large. Indeed, considering the definition of \emph{certifiable robust accuracy} and the certification radius of \Cref{thm:certification}, one can note that
\begin{equation}
  \mathrm{CRA}(\varepsilon)
  = \mathbb{P}_{(x,y)\sim\mathcal{D}}\bigl[\, M(x,y) > L\,\sqrt{2}\,\varepsilon \,\bigr].
  \label{eq:cra}
\end{equation}
This justifies the usage of the hinge loss $\mathcal{H}(x,y) = [L\,\sqrt{2}\varepsilon-M(x,y)]_+$, which pushes the model to produce the highest score for the correct class, ensuring that the difference between the top-2 components is large enough~\cite{anil2019sorting, trockman2021orthogonalizing, brau2023robust}. Lipschitz Cross-entropy proposed by Tsuzuku~et~al.~\cite{tsuzuku2018lipschitz} represents a smooth version of it, extended by Prach~and~Lampert~\cite{prach2022almost} to include a temperature parameter and achieve higher CRA for large radius, which results in the following loss function
\begin{equation}
  \ell(x,y)
  = \mathrm{CrossEntropy}\!\left(
      \mathrm{SoftMax}\left(\frac{F(x) - m\, e_y}{\,t\,}\right),\; y
    \right)t,
  \label{eq:loss_template}
\end{equation}
where $e_y$ is the one-hot vector of the true class, $m\ge 0$ is an {offset} subtracted from the true-class score (sometimes called  \emph{margin}), and $t>0$ is the \emph{temperature}. 

The offset $m$ allows tuning the required certification strength; a typical value is $m=L\,\sqrt{2}\varepsilon$. The temperature provides a smoothing effect, as already observed for learning distillation~\cite{hinton2015distilling}, and larger values helps the CRA for larger radius $\varepsilon$, at the cost of a stronger drop in clean accuracy of the model.
 
A second line of work targets \eqref{eq:cra}, such as GloRo~\cite{leino2021globally}, appends to the output vector of $C$ scores an extra (abstention) class $F_\bot(x) = \max_{j\neq y} F_j(x) + L\,\sqrt{2}\,\varepsilon$ (where $y$ is the top-class), and applies plain cross-entropy to the $C+1$ outputs. A point is certified at radius $\varepsilon$ exactly when the augmented classifier does not select $\bot$, so certification is moved into classification, and the surrogate is differentiable end-to-end. EMMA~\cite{hu2023unlocking} extends the idea, making the offset $m\in\mathbb{R}^C$ depend on the runner-up class of the current sample, so that the large radius is required against the competitor class that is actually close rather than uniformly against all classes. Finally, the logit-annealing loss of BRONet~\cite{lai2025bronet} damps the margin on samples whose margin already comfortably exceeds the target, preventing the optimizer from spending capacity on points that are certified anyway and reallocating it to the bulk. 
\subsection*{The Role of Data Augmentation}
\label{sec:training:data}
Recent advances in image generation via denoising diffusion probabilistic models allow for effective augmentation.  Bartoldson~et~al.~\cite{bartoldson2024adversarial} include generated images in the (adversarial) training process, achieving the current state of the art for empirical robustness. The same boost in performance has been observed in the training of Lipschitz-bounded models~\cite{hu2023unlocking, hu2024recipe, lai2025bronet, hu2026lipnext}.
The current empirical state of the art on CIFAR-10 is trained on $300$M unique diffusion-generated images, for more than $10^{21}$ training FLOPs, to reach $74\%$ robust (yet not certified) accuracy, and the scaling law fitted in the same work predicts that closing the residual gap to its $\sim\!90\%$ plateau would cost on the order of $10^{30}$ FLOPs~\cite{bartoldson2024adversarial}. Training of Lipschitz-bounded models benefits from generated data as well~\cite{altstidl2024scalability}, with certified accuracy improving monotonically in the number of generated samples, adding up to $+8\%$ CRA in some cases~\cite{hu2024recipe}. Furthermore, Prach and Lampert~\cite{prach2025intriguing} measured this dependence on CIFAR-10 and found it logarithmic, i.e., doubling the training set buys about $5\%$ of certified robust accuracy. Data augmentation is therefore an effective ingredient, but likely not one that can be scaled indefinitely.

\section*{Extensions and Open Questions}
\label{sec:extensions}

How are Lipschitz-bounded models linked with other fields? The Lipschitz constant has been treated so far as an instrument for certification; in the rest of the article we show that certification is only one of its uses, and that the same constraint appears in other areas of signal processing.

\subsection*{Implications on Approximation Theory}
\label{sec:extensions:approximation}

Constraining the Lipschitz constant restricts the hypothesis space, so it is natural to ask how much expressivity is paid for it: can a $1$-Lipschitz architecture approximate {every} $1$-Lipschitz function, or only a strict subset? The reference instrument here is the \emph{restricted Stone--Weierstrass} lemma of Anil~et~al.~\cite[Lemma~1]{anil2019sorting}.
Informally, let $\mathcal{X}$ be a compact metric space and let $\mathcal{A}\subseteq C_L(\mathcal{X})$ be a family of $L$-Lipschitz functions that is a \emph{lattice}, i.e., closed under pointwise $\max$ and $\min$. If $\mathcal{A}$ separates points in the strong sense that, for every $x\neq y$ and every pair of values $a,b$ {compatible with the constraint}, i.e., $|a-b|\le L\,d_{\mathcal{X}}(x,y)$, there is some $g\in\mathcal{A}$ with $g(x)=a$ and $g(y)=b$, then $\mathcal{A}$ is dense in $C_L(\mathcal{X})$:
\begin{equation}
  \forall f\in C_L(\mathcal{X}),\ \forall \varepsilon>0,\
  \exists g\in\mathcal{A}\quad:\quad \|f-g\|_\infty<\varepsilon,
\end{equation}
where $\|\cdot\|_\infty$ is the supremum norm on $\mathcal{X}$. The restriction on the admissible values of $a,b$ (also known as the \emph{separability property}) is what separates this statement from the classical version of the theorem. 

The Theorem motivates the design of a dedicated activation, \emph{GroupSort}, which partitions and groups the pre-activations into groups of fixed size and sorts each group in descending order. Being a permutation, GroupSort is norm-preserving and hence exactly $1$-Lipschitz; unlike ReLU, it does not attenuate the gradient norm, which is the property that makes the lattice structure available. Anil~et~al. proved that norm-constrained GroupSort networks form such a lattice and are therefore universal approximators of Lipschitz functions. We note that the result is established in the $\ell_\infty$ geometry, and its transfer to the $\ell_2$ setting used for certification in \Cref{thm:certification} is not immediate.

Read in isolation, this may suggest that ReLU is structurally unsuited to Lipschitz-bounded Networks, and the suspicion is more than heuristic, since norm-constrained ReLU stacks cannot represent the absolute value while preserving the gradient norm~\cite{anil2019sorting}. Nevertheless, Murari~et~al.~\cite[Theorem~3.1]{murari2025approximation} show that $1$-Lipschitz \emph{residual} networks with ReLU activations recover the density result, and hence approximate any $1$-Lipschitz function on a compact set. This closes the loop with the residual constructions of Table~\ref{tab:param_maps} (CPL, SLL), providing evidence that the residual term is not merely convenient for optimization, as observed empirically in~\cite{prach2024lipschitz}, but sufficient for expressivity. It is worth highlighting that universal approximation of Lipschitz functions has been established only for scalar functions, with one-dimensional output, while the general vectorial case is, to the best of our knowledge, unknown.

\subsection*{Randomized Smoothing is Lipschitz}
\label{sec:extensions:smoothing}
Randomized smoothing appears to act at a different stage than the constructions of this article, wrapping an arbitrary base model at test time rather than constraining the architecture. However, the two strategies are intimately related. Let $f:\mathbb{R}^d\to[0,1]$ be measurable function and let $\hat{f}=f*\mathcal{N}(0,\sigma^2 I)$ be its Gaussian smoothing. Denoting by $\Psi$ the standard Gaussian cumulative distribution function, \cite[Lemma~2]{salman2019provably} shows that
\begin{equation}
  x \;\mapsto\; \Psi^{-1}\!\big(\hat{f}(x)\big)
  \quad\text{is } \frac{1}{\sigma}\text{-Lipschitz},
  \label{eq:probit-lipschitz}
\end{equation}
with no regularity required of $f$ beyond measurability, and with a constant that depends on the noise level alone rather than on the base model. Read in the probit coordinate $\Psi^{-1}$, smoothing enforces a Lipschitz constraint, obtained by convolution instead of by construction. Writing $G_i:=\Psi^{-1}\!\circ\hat{F_i}$, the resulting $G$ is Lipschitz-bounded in each output component, and every pair difference $G_i-G_j$ is then $\frac{2}{\sigma}$-Lipschitz. Letting $p_A,p_B$ be the smoothed probabilities of the top two classes at $x$,  applied to $G$ returns
\begin{equation}
  C_g(x)=\frac{G_A(x)-G_B(x)}{2/\sigma}
      =\frac{\sigma}{2}\Big(\Psi^{-1}(p_A)-\Psi^{-1}(p_B)\Big),
\end{equation}
which is exactly the radius proposed by Cohen~et~al.~\cite{cohen2019certified}. To summarize, smoothing provides certification starting from any black-box model, producing a classifier $\hat f$ that has no closed form and is accessible only through Monte Carlo. This is the motivation why the certificate is probabilistic rather than deterministic, where inference cost is proportional to the number sampled noise. 

\subsection*{Lipschitz-based robustness on other Domains}
\label{sec:extensions:domains}
Semantic segmentation is a closely related domain in which Lipschitz-bounded networks have recently found effective applications. In this setting, the input is an image, while the model assigns a class to each output pixel, producing a pixel-wise segmentation. Lipschitz-bounded segmentation models have recently been explored by \cite{massena2026fast} for reliable medical image segmentation, where robustness is particularly important.
However, image processing is not the only field where the idea is applicable. \Cref{thm:certification} is stated for the $\ell_p$ distance on image space, but nothing in its derivation needs a norm $\ell_p$ and an input domain made of images. Abad~Rocamora~et~al.~\cite{rocamora2025certified} take a first step in this direction, bringing Lipschitz-based certification to natural
language processing, where inputs are discrete and of variable length and a dedicated distance is therefore needed. Their construction rests on an extension of the \emph{Levenshtein distance} (itself an extension of the \emph{Hamming} distance) with respect to which the Lipschitz constant of each layer can be bounded. Following the process described in this tutorial, a single forward pass then certifies a radius, which is however, expressed as a number of edit operations on the input tokens, rather than as a continuous $\varepsilon$.

\subsection*{Certificate Soundness}
The guarantees discussed so far are theorems about real numbers, whereas a deployed certifier is a program operating on floating-point weights, and the gap between the two is gaining the attention of the community. Tobler~et~al.~\cite{tobler2025verified} formally verified a certifier for globally-robust networks in Dafny\footnote{\url{https://dafny.org/}}, uncovering exploitable unsoundnesses in the reference implementation of GloRo~\cite{leino2021globally}, severe enough that one of their demonstrations reports a model certified robust on $98.42\%$ of the MNIST test set at $\varepsilon = 1.58$, while a greater ratio of adversarial examples are found within this tolerance. Beyond the individual flaws lies a design-level concern that touches several of the strategies discussed above. The power method converges to the spectral norm from below, so truncating it returns an underestimate of the Lipschitz constant, and since the certificate divides by $L$, an underestimate inflates the radius and yields a false sense of security; bounding from above instead, as Gram iteration~\cite{delattre2023efficient} does by iterating on the matrix rather than on a vector, removes the issue. The concern extends to the numerics of training itself: LipNeXt~\cite{hu2026lipnext} exploits reduced precision to cut the cost of maintaining the constraint, so that the validity of the constant comes to depend on the arithmetic of the whole pipeline rather than of the certification step alone.

\subsection*{Lipschitz Networks Beyond Robustness}
\label{sec:lipnets:beyond}

Certification is not the only reason Lipschitz-bounded models are gaining popularity. In Wasserstein GANs~\cite{arjovsky2017wasserstein} the
Kantorovich--Rubinstein duality allows rewriting the distance $W_1(\mu,\nu)$ between two distributions $\mu,\nu$ as the supremum of $\mathbb{E}_{\mu}[F]-\mathbb{E}_{\nu}[F]$ over $F$ satisfying the $1$-Lipschitz constraint. Lunz~et~al.~\cite{lunz2018adversarial} train a $1$-Lipschitz network $\Psi_\Theta$ to separate ground-truth images from unregularized pseudo-inverse reconstructions, and then reconstruct by minimizing $\|Ax-y\|_2^2+\lambda\Psi_\Theta(x)$. Under a data manifold assumption, the maximizer of the dual is the distance function to the manifold, so the trained regularizer inherits unit gradient norm, i.e., the eikonal condition already met for signed-distance classifiers~\cite{brau2023robust,bethune2023robust}.
In virtual analog audio, the constraint provides convergence guarantees rather than robustness. Massi~et~al.~\cite{massi2024wave} model the nonlinear elements of a circuit with data-driven wave-digital blocks. The iteration converges only if the composed map is a contraction, a condition an unconstrained network cannot be trusted to meet, and which a Lipschitz-bounded network with parameterized spectral layers satisfies by construction. 
In conclusion, for differential privacy, one of the strategies to increase the privacy level consists in ensuring a bound on the gradient norm to calibrate the noise that hides each sample.
B\'ethune~et~al.~\cite{bethune2024dp} obtain that bound from the architecture by constraining the Lipschitz constant of the network with respect to its parameters rather than its input. The bound is then analytical, layer by layer, and known before any data is seen.


\section*{Conclusion}
\label{sec:conclusion}


This tutorial followed the idea behind Lipschitz-based certification, in which a model whose Lipschitz constant is known is certifiable by-design at the cost of one forward pass. The problem reduces to two tasks. Building networks with a known Lipschitz constant is a design problem, answered by where the constraint is imposed, in the loss, in the network, or in the training algorithm; training them to produce large margins is a training problem, answered by losses that act as allocation policies for a fixed margin budget.

%
\bibliographystyle{IEEEtran}
\bibliography{references}

%
%
\section*{Biographies}
\label{sec:bio}

\begin{IEEEbiographynophoto}{Fabio Brau} is an Assistant Professor at the University of Cagliari, conducting research on AI security and trustworthiness. He holds a Master's in Mathematics from the University of Pisa and a Ph.D. in Embedded Systems from the Scuola Superiore Sant'Anna, where he worked on robustness certification for deep neural networks. His research spans Adversarial Machine Learning and Trustworthy AI for Computer Vision and NLP systems, with contributions to Lipschitz-constrained architectures for certifiable robustness. He has published in top-tier venues including NeurIPS, CVPR, AAAI, and IEEE TPAMI, and is involved in European research projects on trustworthy AI.
\end{IEEEbiographynophoto}
\vspace{-10pt} 
\begin{IEEEbiographynophoto}{Giorgio Piras} is a Postdoctoral Researcher at the sAIfer Lab, Department of Electrical and Electronic Engineering, University of Cagliari, Italy, working on adversarial machine learning and large language model security. He received his BSc (110/110, 2019) and MSc with honors (2021) from the University of Cagliari, and his PhD with honors (2025) from Sapienza University of Rome (National PhD-AI program), with the thesis \emph{Adversarial Pruning: Improving Evaluations and Methods}, supervised by Prof. Battista Biggio. He was a visiting student at Karlsruhe Institute of Technology. He reviews for NeurIPS, ICLR, AAAI, USENIX, ACM CCS, Pattern Recognition, Neurocomputing, Machine Learning, and IEEE TIFS.
\end{IEEEbiographynophoto}
\vspace{-10pt} 
\begin{IEEEbiographynophoto}{Maura Pintor}
(Member, IEEE) is an Assistant Professor at the PRA Lab, Department of Electrical and Electronic Engineering, University of Cagliari, Italy, where she received her PhD in Electronic and Computer Engineering in 2022. Her research focuses on adversarial machine learning, particularly evaluating, debugging, and improving the robustness of machine learning systems to make AI more reliable, secure, and trustworthy. She is an Area Chair for NeurIPS, Associate Editor for Pattern Recognition, and Consulting Associate Editor for IEEE TIFS. She is a member of the IEEE Information Forensics and Security Technical Committee, IAPR, and ELLIS, and maintains the open-source library SecML-Torch.
\end{IEEEbiographynophoto}
%
\begin{IEEEbiographynophoto}{Battista Biggio}
(Fellow, IEEE) (MSc 2006, PhD 2010) is Full Professor at the University of Cagliari, Italy. He has provided pioneering contributions in machine learning security, playing a leading role in this field. His seminal paper on ``Poisoning Attacks against Support Vector Machines'' won the prestigious 2022 ICML Test of Time Award. His work on ``Wild Patterns'' won the 2021 Best Paper Award and Pattern Recognition Medal from Elsevier Pattern Recognition. He has managed more than 10 research projects, and serves as a PC member of ICML and USENIX Security, and as Area Chair of NeurIPS. He chaired IAPR TC1 (2016-2020), and served as Associate Editor for IEEE TNNLS, IEEE CIM, and Elsevier PRJ. He is now Associate Editor-in-Chief for PRJ. He is also Fellow of IEEE, Senior Member of ACM, and member of IAPR and ELLIS. 
\end{IEEEbiographynophoto}

\vfill

\end{document}

%% file: commons-utils.tex
\newcommand{\thetitle}{Lipschitz Thinking: \add{Ten Years of Certifiable-by-Design Robust Neural Networks}}

\newcommand{\cmark}{\ding{51}} 
\newcommand{\xmark}{\ding{55}} 

\newcommand{\add}[1]{{#1}}
\newcommand{\del}[1]{}
